\documentclass{bmcart}

\usepackage{amsthm,amsmath}
\usepackage[utf8]{inputenc} 

\usepackage[misc]{ifsym}
\usepackage[colorinlistoftodos]{todonotes}
\usepackage{graphicx}

\usepackage{calc}

\usepackage{multirow}
\usepackage{rotating}

\makeatletter
\def\Hline{%
\noalign{\ifnum0=`}\fi\hrule \@height 1pt \futurelet
\reserved@a\@xhline}
\makeatother

\startlocaldefs
\endlocaldefs

\begin{document}

\begin{frontmatter}

\begin{fmbox}
\dochead{Research}


\title{MADGAN: unsupervised Medical Anomaly Detection GAN using multiple adjacent brain MRI slice reconstruction}


\author[
   addressref={LP},                   
   corref={LP},                       
   email={han@lpixel.net}   
]{\inits{CH}\fnm{Changhee} \snm{Han}}
\author[
   addressref={CamRadiol,CRUK},
   email={lr495@cam.ac.uk}
]{\inits{LR}\fnm{Leonardo} \snm{Rundo}}
\author[
   addressref={NII},
  email={k-murao@nii.ac.jp}
]{\inits{KM}\fnm{Kohei} \snm{Murao}}
\author[
   addressref={NCGM},
  email={tnogucci@radiol.med.kyushu-u.ac.jp}
]{\inits{TM}\fnm{Tomoyuki} \snm{Noguchi}}
\author[
   addressref={LP},
  email={shimahara@lpixel.net }
]{\inits{YS}\fnm{Yuki} \snm{Shimahara}}
\author[
   addressref={ELTE},
  email={zoltanmilacski@gmail.com}
]{\inits{ZAM}\fnm{\\Zolt\'{a}n \'{A}d\'{a}m} \snm{Milacski}}
\author[
   addressref={JU},
  email={s-koshino@juntendo.ac.jp}
]{\inits{SK}\fnm{Saori} \snm{Koshino}}
\author[
   addressref={CamRadiol,CRUK},
   email={es220@medschl.cam.ac.uk}
]{\inits{ES}\fnm{Evis} \snm{Sala}}
\author[
   addressref={UTokyo,WPI},
  email={nakayama@ci.i.u-tokyo.ac.jp}
]{\inits{HN}\fnm{Hideki} \snm{Nakayama}}
\author[
   addressref={NII},
  email={satoh@nii.ac.jp}
]{\inits{SS}\fnm{Shin'ichi} \snm{Satoh}}


\address[id=LP]{
  \orgname{LPIXEL Inc.}, 
  \city{Tokyo},                              
  \cny{Japan}                                    
}
\address[id=CamRadiol]{%
  \orgname{Department of Radiology, University of Cambridge},
  \city{Cambridge},
  \cny{United Kingdom}
}
\address[id=CRUK]{%
  \orgname{Cancer Research UK Cambridge Centre, University of Cambridge},
  \city{Cambridge},
  \cny{United Kingdom}
}
\address[id=NII]{
  \orgname{Research Center for Medical Big Data, National Institute of Informatics}, 
  \city{Tokyo},                              
  \cny{Japan}                                    
}
\address[id=NCGM]{
  \orgname{National  Center  for  Global  Health  and  Medicine}, 
  \city{Tokyo},                              
  \cny{Japan}                                    
}
\address[id=JU]{
  \orgname{Department of Radiology, Juntendo University}, 
  \city{Tokyo},                              
  \cny{Japan}                                    
}
\address[id=ELTE]{%
  \orgname{Department of Artificial Intelligence, ELTE E\"{o}tv\"{o}s Lor\'{a}nd University},
  \city{Budapest},
  \cny{Hungary}
}
\address[id=UTokyo]{
  \orgname{Graduate School of Information Science and Technology, The University of Tokyo}, 
  \city{Tokyo},                              
  \cny{Japan}                                    
}
\address[id=WPI]{
  \orgname{International Research Center for Neurointelligence (WPI-IRCN), The University of Tokyo Institutes for Advanced Study, The University of Tokyo},
  \city{Tokyo},                          
  \cny{Japan}                            
}


\begin{artnotes}
\end{artnotes}

\end{fmbox}


\begin{abstractbox}

\begin{abstract} 
\parttitle{Background} Unsupervised learning can discover various unseen abnormalities, relying on large-scale unannotated medical images of healthy subjects. Towards this, unsupervised methods reconstruct a 2D/3D single medical image to detect outliers either in the learned feature space or from high reconstruction loss. However, without considering continuity between multiple adjacent slices, they cannot directly discriminate diseases composed of the accumulation of subtle anatomical anomalies, such as Alzheimer's Disease (AD). Moreover, no study has shown how unsupervised anomaly detection is associated with either disease stages, various (i.e., more than two types of) diseases, or multi-sequence  Magnetic Resonance Imaging (MRI) scans.

\parttitle{Results} We propose unsupervised Medical Anomaly Detection Generative Adversarial Network (MADGAN), a novel two-step method using GAN-based multiple adjacent brain MRI slice reconstruction to detect brain anomalies at different stages on multi-sequence structural MRI: (\textit{Reconstruction}) Wasserstein loss with Gradient Penalty + $100$ $\ell _1$ loss---trained on $3$ healthy brain axial MRI slices to reconstruct the next $3$ ones---reconstructs unseen healthy/abnormal 
scans; (\textit{Diagnosis}) Average $\ell _2$ loss per scan discriminates them, comparing the ground truth/reconstructed slices. For training, we use two different datasets composed of $1,133$ healthy T1-weighted (T1) and $135$ healthy contrast-enhanced T1 (T1c) brain MRI scans for detecting AD and brain metastases/various diseases, respectively. Our Self-Attention MADGAN can detect AD on T1 scans at a very early stage, Mild Cognitive Impairment (MCI), with Area Under the Curve (AUC) $0.727$, and AD at a late stage with AUC $0.894$, while detecting brain metastases on T1c scans with AUC $0.921$.

\parttitle{Conclusions}
Similar to physicians' way of performing a diagnosis, using massive healthy training data, our first multiple MRI slice reconstruction approach, MADGAN, can reliably predict the next $3$ slices from the previous $3$ ones only for unseen healthy images. As the first unsupervised various disease diagnosis, MADGAN can reliably detect the accumulation of subtle anatomical anomalies and hyper-intense enhancing lesions, such as (especially late-stage) AD and brain metastases on multi-sequence MRI scans.

\end{abstract}







\end{abstractbox}
%

\end{frontmatter}



\section*{Background}
\label{sec:background}
Machine Learning has revolutionized life science research, especially in Neuroimaging and Bioinformatics~\cite{gao2017,serra2018}, such as by modeling interactions between whole brain genomics/imaging~\cite{park2012,medland2014} and identifying Alzheimer's Disease (AD)-related proteins~\cite{zhao2019}. Especially, Deep Learning can achieve accurate computer-assisted diagnosis when large-scale annotated training samples are available. In Medical Imaging, unfortunately, preparing such massive annotated datasets is often unfeasible \cite{han2020AIAI,cheplygina2019}; to tackle this pervasive problem, researchers have proposed various data augmentation techniques, including Generative Adversarial Network (GAN)-based ones~\cite{goodfellow2014,FridAdar,Han2020WIRN,Han2,Han3,han2019CIKM}
; alternatively, Rauschecker \textit{et al.} combined Convolutional Neural Networks (CNNs), feature engineering, and expert-knowledge Bayesian network to derive brain  Magnetic Resonance Imaging (MRI) differential diagnoses that approach neuroradiologists' accuracy for $19$ diseases. However, even exploiting these techniques, supervised learning still requires many images with pathological features, even for rare disease, to make a reliable diagnosis; nevertheless, it can only detect already-learned specific pathologies. In this regard, as physicians notice previously unseen anomaly examples using prior information on healthy body structure, unsupervised anomaly detection methods leveraging only large-scale healthy images can discover and alert overlooked diseases when their generalization fails.

Towards this, researchers reconstructed a single medical image \textit{via} GANs~\cite{Schlegl}, AutoEncoders (AEs)~\cite{Uzunova}, or combining them, since GANs can generate realistic images and AEs, especially Variational AEs (VAEs), can directly map data onto its latent representation~\cite{Chen}; then, unseen images were scored by comparing them with reconstructed ones to discriminate a pathological image distribution (i.e., outliers either in the learned feature space or from high reconstruction loss). However, those single image reconstruction methods mainly target diseases easy-to-detect from a single image even for non-expert human observers, such as glioblastoma on MR images~\cite{Chen} and lung cancer on Computed Tomography 
(CT) images~\cite{Uzunova}. Without considering continuity between multiple adjacent images, they cannot directly discriminate diseases composed of the accumulation of subtle anatomical anomalies, such as AD. Moreover, no study has shown so far how unsupervised anomaly detection is associated with either disease stages, various (i.e., more than $2$ types of) diseases, or multi-sequence MRI scans.

Therefore, this paper proposes unsupervised Medical Anomaly Detection GAN (MADGAN), a novel two-step method using GAN-based multiple adjacent brain MRI slice reconstruction to detect various diseases at various stages on multi-sequence structural MRI (Fig.~\ref{fig1}): (\textit{Reconstruction}) Wasserstein loss with Gradient Penalty (WGAN-GP)~\cite{Gulrajani,han2018} + 100 $\ell _1$ loss---trained on $3$ healthy brain axial MRI slices to reconstruct the next $3$ ones---reconstructs unseen healthy/abnormal scans; the $\ell _1$ loss generalizes well only for unseen images with a similar distribution to the training images while the WGAN-GP loss captures recognizable structure; (\textit{Diagnosis}) Average $\ell _2$ loss per scan discriminates them, comparing the ground truth/reconstructed slices; the $\ell _2$ loss clearly discriminates the healthy/abnormal scans as squared error becomes huge for outliers. Using Receiver Operating Characteristics (ROCs) and their Area Under the Curves (AUCs), we evaluate the diagnosis performance of AD on T1-weighted (T1) MRI scans, and brain metastases/various diseases (e.g., small infarctions, aneurysms) on contrast-enhanced T1 (T1c) MRI scans. Using $1,133$ healthy T1 and $135$ healthy T1c scans for training, our Self-Attention (SA) MADGAN approach can detect AD at a very early stage, Mild Cognitive Impairment (MCI), with AUC $0.727$, and AD at a late stage with AUC $0.894$, while detecting brain metastases with AUC $0.921$.



\paragraph{Contributions.} Our main contributions are as follows:
\begin{itemize}
\item \textbf{MRI Slice Reconstruction:} This first multiple MRI slice reconstruction approach can reliably predict the next $3$ slices from the previous $3$ ones only for unseen images similar to training data by combining SAGAN and $\ell _1$ loss.

\item \textbf{Unsupervised Anomaly Detection:} This first unsupervised multi-stage anomaly detection reveals that, like physicians' way of performing a diagnosis, massive healthy data can aid early diagnosis, such as of MCI, while also detecting late-stage disease much more accurately by discriminating with $\ell _2$ loss.

\item \textbf{Various Disease Diagnosis:} This first unsupervised various disease diagnosis can reliably detect the accumulation of subtle anatomical anomalies (e.g., AD), as well as  hyper-intense enhancing lesions (e.g., brain metastases) on multi-sequence MRI scans.



\end{itemize}

\section*{Related work}
\label{sec:RelatedWork}

\subsection*{Alzheimer's disease diagnosis}
\label{sec:ADdiagnosis}

Even though the clinical, social, and economic impact of early AD diagnosis is of paramount importance \cite{arvanitakis2019diagnosis}---primarily associated with MCI detection \cite{moscoso2019prediction}---it generally relies on subjective assessment by physicians (e.g., neurologists, geriatricians, and psychiatrists).
The diagnosis typically considers two characteristics: (\textit{i}) mesial temporal lobe atrophy (particularly hippocampus, entorhinal cortex, and perirhinal cortex) and (\textit{ii}) temporo-parietal cortical atrophy. Quantifying these structures is crucial for early AD diagnosis and its progression tracking \cite{desikan2009}. Moreover, morphometry-based markers, such as gray matter volume and cortical thickness, can play a key role in brain atrophy assessment \cite{ma2016}.

Towards quantitative and reproducible approaches, many  traditional supervised Machine Learning-based methods---which relies on handcrafted MRI-derived features---were proposed in the literature~\cite{salvatore2015,nanni2019}.
In this context, diffusion-weighted MRI tractography enables reconstructing the brain's physical connections that can be subsequently investigated by complex network-based techniques.
Lella \textit{et al.} \cite{lella2020} employed the whole brain structural communicability as a graph-based metric to describe the AD-relevant brain connectivity disruption.
This approach achieved comparable performance with classic Machine Learning models---namely, Support Vector Machines, Random Forests, and Artificial Neural Networks---in terms of classification and feature importance analysis.

In the latest years, Deep Learning has achieved outstanding performance by exploiting more multiple levels of abstraction and descriptive embeddings in a hierarchy of increasingly complex features \cite{lecun2015}: Liu \textit{et al.} devised a semi-supervised CNN to significantly reduce the need for labeled training data~\cite{liu2014early}; for clinical decision-making tasks, Suk \textit{et al.} integrated multiple sparse regression models (i.e., Deep Ensemble Sparse Regression Network)~\cite{suk2017}; Spasov \textit{et al.} proposed a parameter-efficient CNN for 3D separable convolutions, combining dual learning and a specific layer to predict the conversion from MCI to AD within $3$ years~\cite{spasov2019}; different from CNN-based approaches, Parisot used a semi-supervised Graph Convolutional Network trained on a sub-set of labeled nodes with diagnostic outcomes to represent sparse clinical data~\cite{parisot2018}. However, to the best of our knowledge, no existing work has conducted fully unsupervised anomaly detection for AD diagnosis since capturing subtle anatomical differences between MCI and AD is challenging.

\subsection*{Brain metastasis and various disease diagnosis}
\label{sec:AbnormalityMRIdiagnosis}

Along with neuro-degenerative diseases, MRI can also play a definite role in abnormality diagnosis. Whereas advanced cancer screening, imaging, and therapeutics can improve oncological patients' survival and quality of life, brain metastases still remain major contributors of morbidity and mortality, especially for patients with lung cancer, breast cancer, or malignant melanoma~\cite{sacks2020}. To tackle this, previous computational methods have detected the brain metastases in either supervised~\cite{han2019CIKM,grovik2020} or semi-automatic manner~\cite{rundo2018NC,rundo2018next}.

Detecting other various diseases, such as cerebral aneurysms, hemorrhage, and infarctions, also remain challenging~\cite{miki2016,vert2017}. Therefore, similar to the brain metastases, researchers have mostly relied on supervised methods, especially CNN-based detection~\cite{zhou2019,conti2020,federau2020}. Recently, unsupervised anomaly segmentation methods have been applied to brain MRI datasets for detecting multiple sclerosis lesions~\cite{baur2018deep} and glioblastoma~\cite{zimmerer2019unsupervised}. However, it is difficult to directly compare our approach with such existing unsupervised anomaly detection methods on 3D medical images since we perform a whole-brain diagnosis (i.e., classification), instead of segmentation.

\subsection*{Unsupervised medical anomaly detection}
Unsupervised disease diagnosis is challenging because it requires estimating healthy anatomy's normative distributions only from healthy examples to detect outliers either in the learned feature space or from high reconstruction loss. The latest advances in Deep Learning, mostly GANs~\cite{goodfellow2014} and VAEs~\cite{kingma2013}, have allowed for the accurate estimation of the high-dimensional healthy distributions. Except for discriminative-boundary-based approaches including \cite{alaverdyan2020regularized}, 
almost all unsupervised medical anomaly detection studies have leveraged reconstruction: as pioneering research, Schlegl
\textit{et al.} proposed AnoGAN to detect outliers in the learned feature space of the GAN~\cite{schlegl2017unsupervised}; then, the same authors presented fast AnoGAN that can efficiently map query images onto the latent space~\cite{Schlegl}; since the reconstruction-based models often suffer from many false positives, Chen \textit{et al.} penalized large deviations between original/reconstructed images in gliomas and stroke lesion detection on brain MRI~\cite{chen2020}. However, to the best of our knowledge, all previous studies are based on 2D/3D single image reconstruction, without considering continuity between multiple adjacent slices. Moreover, no existing work has investigated how unsupervised anomaly detection is associated with either disease stages, various (i.e., more than two types of) diseases, or multi-sequence MRI scans.


\subsection*{Self-Attention GANs (SAGANs)}
Zhang \textit{et al.} proposed SAGAN that deploys an SA mechanism in the generator/discriminator of a GAN to learn global and long-range dependencies for diverse image generation~\cite{zhang2019self}; for further performance improvement, they suggested to apply the SA modules to large feature maps. The SAGANs have shown great promise in various tasks, such as human pose estimation~\cite{wang2019improving}, image colorization~\cite{sharma2019robust}, photo-realistic image de-quantization~\cite{zhang2020deep}, and large-scale image generation~\cite{brock2018large}. This 
SAGAN trend also applies to Medical Imaging to extract multi-level features for better super-resolution/denoising and lesion characterization: to mitigate the problem of thin slice thickness, Kudo \textit{et al}. and Li \textit{et al.} applied the SA modules to GANs on CT and MRI scans, respectively~\cite{kudo2019virtual,li2020super}; similarly, in \cite{lan2020sc}, the authors proposed to fuse plane SA modules and depth SA modules for low-dose 
3D CT denoising; Lan \textit{et al.} synthesized multi-modal 3D brain images using SA conditional GAN~\cite{lan2020sc}; Ali \textit{et al.} incorporated SA modules into progressive growing of GANs to generate realistic and diverse skin lesion images for data augmentation~\cite{ali2019data}. However, to the best of our knowledge, no existing work has directly exploited the SAGAN for medical disease diagnosis.

\section*{Materials and methods}

\subsection*{Datasets}
\label{sec:datasets}

\subsubsection*{AD dataset: OASIS-3}
\label{sec:OASIS} 
We use a longitudinal $3.0$T MRI dataset of $176 \times 240$/$176 \times 256$ T1 brain axial MRI slices containing both normal aging subjects/AD patients, extracted from the Open Access Series of Imaging Studies-3 (OASIS-3)~\cite{lamontagne2018oasis}. The $176 \times 240$ slices are zero-padded to reach $176 \times 256$ pixels. Relying on Clinical Dementia Rating (CDR)~\cite{Morris}, common clinical scale for the staging of dementia, the subjects are comprised of:

\begin{itemize}
\item Unchanged CDR $= 0$: Cognitively healthy population;
\item CDR $= 0.5$: Very mild dementia\ ($\sim$ MCI);
\item CDR $= 1$: Mild dementia;
\item CDR $= 2$: Moderate dementia.
\end{itemize}

Since our dataset is longitudinal and the same subject's CDRs may vary (e.g., CDR $= 0$ to CDR $= 0.5$), we only use scans with unchanged CDR $= 0$ to assure certainly healthy scans. As CDRs are not always assessed simultaneously with the MRI acquisition, we label MRI scans with CDRs at the closest date. We only select brain MRI slices including hippocampus/amygdala/ventricles among whole $256$ axial slices per scan to avoid over-fitting from AD-irrelevant information; the atrophy of the hippocampus/amygdala/cerebral cortex, and enlarged ventricles are strongly associated with AD, and thus they mainly affect the AD classification performance of Machine Learning~\cite{Ledig}. Moreover, we discard low-quality MRI slices. The remaining dataset is divided as follows:

\begin{itemize}
\item Training set: Unchanged CDR $= 0$ ($408$ subjects/$1,133$ scans/$57,834$ slices);
\item Test set: Unchanged CDR $= 0$ ($168$ subjects/$473$ scans/$24,278$ slices),\\CDR $= 0.5$ ($152$ subjects/$253$ scans/$13,813$ slices),\\CDR $= 1$ ($90$ subjects/$135$ scans/$7,532$ slices),\\CDR $= 2$ ($6$ subjects/$10$ scans/$500$ slices).
\end{itemize}
The same subject's scans are included in the same dataset. The datasets are strongly biased towards healthy scans similar to MRI inspection in the clinical routine. During training for reconstruction, we only use the training set---structural MRI alone---containing healthy slices to conduct unsupervised learning. We do not use a validation set as our unsupervised diagnosis step is non-trainable.

\subsubsection*{Brain metastasis and various disease dataset}
\label{sec:dataset2}
This paper also uses a non-longitudinal, heterogeneous $1.5$T/$3.0$T MRI dataset of $190 \times 224$/$216 \times 256$/$256 \times 256$/$460 \times 460$ T1c brain axial MRI slices.
This dataset was collected by the authors at the National Center for Global Health and Medicine, Tokyo, Japan, and is currently not publicly available due to ethical restrictions. The dataset contains both healthy subjects, brain metastasis patients \cite{rundo2018NC}, and patients with various diseases different from brain metastases. The slices are resized to $176 \times 256$ pixels. The various diseases include but are not limited to:

\begin{itemize}
\item Small infarctions;
\item Aneurysms;
\item Benign tumors;
\item Hemorrhages;
\item Cysts;
\item White matter lesions;
\item Post-operative inflammations.
\end{itemize}

Conforming to T1 slices, we also only select T1c slices including hippocampus, amygdala, and ventricles---a large portion of various diseases also appear in the mid-brain. The remaining dataset is divided as follows:

\begin{itemize}
\item Training set: Normal ($135$ subjects/$135$ scans/$7,793$ slices);
\item Test set: Normal ($58$ subjects/$58$ scans/$3,353$ slices),\\Brain Metastases ($79$ subjects/$79$ scans/$4,872$ slices),\\Various Diseases ($66$ subjects/$66$ scans/$4,195$ slices).
\end{itemize}
Since we cannot collect large-scale T1c scans from healthy patients like OASIS-3 dataset, during training for reconstruction, we use both T1/T1c training sets containing healthy slices simultaneously for the knowledge transfer. In the clinical practice, T1c MRI is well-established in detecting various diseases, including brain metastases \cite{arvold2016updates}, thanks to its high-contrast in the enhancing region---however, the contrast agent is not suitable for screening studies. Accordingly, such inter-sequence knowledge transfer is valuable in computer-assisted MRI diagnosis. During testing, we make an unsupervised diagnosis on T1 and T1c scans separately.

\subsection*{MADGAN-based multiple adjacent brain MRI slice reconstruction}
\label{sec:MRIrecon}
To model strong consistency in healthy brain anatomy (Fig.~\ref{fig1}), in each scan, we reconstruct the next $3$ MRI slices from the previous $3$ ones using an image-to-image GAN (e.g., if a scan includes $40$ slices $s_i$ for $i=1,\dots,40$, we reconstruct all possible $35$ setups: $(s_i)_{i\in\{1,2,3\}} \mapsto (s_i)_{i\in\{4,5,6\}}$; $(s_i)_{i\in\{2,3,4\}} \mapsto (s_i)_{i\in\{5,6,7\}}$; \dots; $(s_i)_{i\in\{35,36,37\}} \mapsto (s_i)_{i\in\{38,39,40\}}$).
As Fig.~\ref{fig2} shows, our MADGAN uses a U-Net-like~\cite{Ronneberger,RundoUSEnet} generator with $4$ convolutional layers in encoders and $4$ deconvolutional layers in decoders respectively with skip connections, as well as a discriminator with $3$ decoders.
We apply batch normalization to both convolution with Leaky Rectified Linear Unit (ReLU) and deconvolution with ReLU.
Between the designated convolutional/deconvolutional layers and batch normalization layers, we apply SA modules~\cite{zhang2019self} for effective knowledge transfer $via$ feature recalibration between T1 and T1c slices;
as confirmed on four different image datasets~\cite{kimura2020adversarial}, introducing the SA modules to GAN-based anomaly detection (i.e., attention-driven, long-range dependency modeling) can also mitigate the effect of noise by ignoring irrelevant disturbances and focusing on the salient body parts in the slice.
We compare the MADGAN models with a different number of the SA modules: (\textit{i}) no SA modules (i.e., MADGAN); (\textit{ii}) 3 (red-contoured) SA modules (i.e., 3-SA MADGAN); (\textit{iii}) 7 (red- and blue-contoured) SA modules (i.e., 7-SA MADGAN). To confirm how reconstructed slices' realism and anatomical continuity affect medical anomaly detection, we also compare the MADGAN models with different loss functions: (\textit{i}) WGAN-GP loss + 100 $\ell _1$ loss (i.e., MADGAN); (\textit{ii}) WGAN-GP loss (i.e., MADGAN w/o $\ell _1$ loss). The $\ell _1$ and $\ell _2$ losses between an input image $x$ and its reconstructed image $x'$ are defined as follows:
\begin{align}
&&\ell _1 = \sum_{i=1}^{P} |x_i - x'_i|,\\
&&\ell _2 = \sum_{i=1}^{P} (x - x')^2,
\end{align}
where $P$ denotes the number of pixels.

\paragraph*{Implementation details} Each MADGAN training lasts for $1.8 \times 10^{6}$ steps with a batch size of $16$ (our maximum available batch size). We use $2.0 \times 10^{-4}$ learning rate for Adam optimizer~\cite{kingma2014}. Such as in RGB images, we concatenate adjacent $3$ grayscale slices into $3$ channels. During training, the generator uses two dropout~\cite{srivastava2014dropout} layers with 0.5 rate. We flip the discriminator’s real/synthetic labels once in three times for robustness. Using 4 NVIDIA Quadro GV100 graphics processing units, we implement the framework on TensorFlow 1.8.

\subsection*{Unsupervised medical anomaly detection}
\label{sec:UnsupADdiagnosis}

During diagnosis, we classify unseen healthy and abnormal scans based on average $\ell _2$ loss per scan. The average $\ell _2$ loss is calculated from whole MADGAN-reconstructed $3$ slices $s_i$ of each scan containing $n$ slices: $(s_i)_{i\in\{4,5,6\}}$; $(s_i)_{i\in\{5,6,7\}}$; \dots; $(s_i)_{i\in\{n-2,n-1,n\}}$. We use the $\ell _2$ loss since squared error is sensitive to outliers and it significantly outperformed other losses (i.e., $\ell _1$ loss, Dice loss, Structural Similarity loss) in our preliminary paper~\cite{han2020CIBB}.
To evaluate its unsupervised AD diagnosis performance on a T1 MRI test set, we show ROCs---along with the AUC values---between CDR $= 0$ \textit{vs} (\textit{i}) all the other CDRs; (\textit{ii}) CDR $= 0.5$; (\textit{iii}) CDR $= 1$; (\textit{iv}) CDR $= 2$. We also show the AUCs under different training steps (i.e., $150$k, $300$k, $600$k, $900$k, $1.8$M steps) and confirm the effect of calculating average $\ell _2$ loss (among whole slice sets or continuous 10 slice sets exhibiting the highest loss) per scan; if the 10 slice sets start from the $j$-th slice, we use: $(s_i)_{i\in\{j,j+1,j+2\}}$; $(s_i)_{i\in\{j+1,j+2,j+3\}}$; \dots; $(s_i)_{i\in\{j+9,j+10,j+11\}}$).
Moreover, we visualize pixelwise $\ell _2$ loss between real/reconstructed 3 slices, along with distributions of average $\ell _2$ loss per scan of CDR $= 0/0.5/1/2$ to know how disease stages affect its discrimination. In exactly the same manner, we evaluate the diagnosis performance of brain metastases/various diseases on a T1c MRI test set, showing ROCs/AUCs between normal \textit{vs} (\textit{i}) brain metastases + various diseases; (\textit{ii}) brain metastases; (\textit{iii}) various diseases.

\section*{Results}
\label{sec:Results}

\subsection*{Reconstructed brain MRI slices}
\label{sec:MRIreconRes}
Fig.~\ref{fig3} illustrates example real T1 MRI slices from a test set and their reconstruction by MADGAN and 7-SA MADGAN. Similarly, Figs.~\ref{fig4} and~\ref{fig5} show example real T1c MRI slices and their reconstructions. Pixelwise $\ell _2$ loss tends to increase (i.e., high intensity in the heatmap) around lesions due to their different image distribution from healthy samples.

Figs.~\ref{fig6} and~\ref{fig7} indicate distributions of average $\ell _2$ loss per scan on T1 and T1c scans, respectively. Leveraging $\ell _1$ loss' good realism sacrificing diversity (i.e., generalizing well only for unseen images with a similar distribution to training images) and WGAN-GP loss' ability to capture recognizable structure, the MADGAN can successfully capture T1-specific appearance and anatomical changes from the previous $3$ slices. Meanwhile, the 7-SA MADGAN tends to be less stable in keeping texture but more sensitive to abnormal anatomical changes due to the SA modules' anomaly-sensitive reconstruction $via$ the attention-driven, long-range dependency modeling, resulting in moderately higher average $\ell _2$ loss than the MADGAN.

Since the models are trained only on healthy slices, as visualized by an overimposed Jet colormap, reconstructing slices with higher CDRs tends to comparatively fail, especially around hippocampus, amygdala, cerebral cortex, and ventricles due to their insufficient atrophy after reconstruction; this is plausible because physicians also perform the AD diagnosis based on their prior normal atrophy information around those body parts. We do not find other significant reconstruction failures except them, considering that inter-subject/sequence variability also lead to considerable reconstruction failures. The T1c scans show much lower average $\ell _2$ loss than the T1 scans due to darker texture. Since most training images are the T1 slices with brighter texture than the T1c slices, reconstruction quality clearly decreases on the T1c slices, occasionally exhibiting bright texture. Accordingly, reconstruction failure from anomaly contributes comparatively less to the average $\ell _2$ loss, especially when local small lesions, such as brain abscess and enhanced lesions, appear---unlike global big lesions, such as multiple cerebral infarction and blood component retention. However, 
the average $\ell _2$ loss remarkably increases on brain metastases scans due to their hyper-intensity, especially for the 7-SA MADGAN.

\subsection*{Unsupervised anomaly detection results}
\label{sec:ADdiagnosisRes}
Figs.~\ref{fig8} and~\ref{fig9} show AUCs of unsupervised anomaly detection on T1 and T1c scans under different training steps, respectively.
The AUCs generally increase as training progresses, but more SA modules require more training steps until convergence due to their feature recalibration.
Although most models show a convergence after $900$k steps, MADGAN with abundant SA modules might perform even better, especially on the T1c scans with less training data than the T1 scans, if we continue its training.

All the best results in specific tasks, except for CDR $= 0$ \textit{vs} CDR $= 0.5$, are from 
the SA models (e.g., 7-SA MADGAN w/o $\ell _1$ loss under 900k steps: AUC $0.783$ in CDR $= 0$ \textit{vs} CDR $= 0.5 + 1 + 2$, 3-SA MADGAN under 300k steps: AUC $0.966$ in normal \textit{vs} brain metastases, 3-SA MADGAN under 600k steps: AUC $0.638$ in normal \textit{vs} various diseases); thus, whereas the SA models, which do not know the task to optimize in an unsupervised manner, perform unstably, we might use them similarly 
to supervised learning if we could obtain good parameters for a certain disease. Without $\ell _1$ loss, the AUCs tend to decrease, also accompanying large fluctuations; 7-SA MADGAN w/o $\ell _1$ loss performs well on the T1 scans but poorly on the T1c scans due to the instability.

Figs.~\ref{fig10} and~\ref{fig11} illustrate ROC curves and their AUCs on T1 and T1c scans under $1.8$M training steps, respectively. Since brains with higher CDRs accompany stronger anatomical atrophy from healthy brains, their AUCs between unchanged CDR $= 0$ remarkably increase as CDRs increase. MADGAN and 7-SA MADGAN both achieves good AUCs, especially for higher CDRs---The MADGAN obtains AUC $0.750/0.707/0.829$ in CDR $= 0$ \textit{vs} CDR $= 0.5/1/2$, respectively; the discrimination between healthy subjects \textit{vs} MCI patients (i.e., CDR $= 0$ \textit{vs} CDR $= 0.5$) is extremely difficult even in a supervised manner~\cite{Ledig}. Whereas detecting various diseases is difficult in an unsupervised manner, the 7-SA MADGAN outperforms the MADGAN and achieves AUC $0.921$ in brain metastases detection. As Tables~\ref{tab:CDR} and~\ref{tab:T1c} show, the effect of 
how to calculate average $\ell _2$ loss (among whole slice sets or continuous 10 slice sets exhibiting the highest loss) per scan is limited. Whereas no significant differences exist between them, the best performing approach on each dataset is always whole slice sets-based.

\section*{Discussion and conclusions}
\label{sec:Conclusion}


Using massive healthy data, our MADGAN-based multiple MRI slice reconstruction can reliably discriminate AD patients from healthy subjects for the first time in an unsupervised manner; to detect the accumulation of subtle anatomical anomalies, our solution leverages a two-step approach: (\textit{Reconstruction}) $\ell _1$ loss generalizes well only for unseen images with a similar distribution to training images while WGAN-GP loss captures recognizable structure; (\textit{Diagnosis}) $\ell _2$ loss clearly discriminates healthy/abnormal data as squared error becomes huge for outliers. Using $1,133$ healthy T1 MRI scans for training, our approach can detect AD at a very early stage, MCI, with AUC $0.727$ while detecting AD at a late stage with AUC $0.894$. Accordingly, this first unsupervised anomaly detection across different disease stages reveals that, like physicians' way of performing a diagnosis, large-scale healthy data can reliably aid early diagnosis, such as of MCI, while also detecting late-stage disease much more accurately.

To confirm its ability to also detect other various diseases, even on different MRI sequence scans, we firstly investigate how unsupervised medical anomaly detection is associated with various diseases and multi-sequence MRI scans, respectively. Due to the different texture of T1/T1c slices, reconstruction quality clearly decreases on the data-sparse T1c slices, and thus reconstruction failure from anomaly contributes comparatively less to the average $\ell _2$ loss. Nevertheless, we generally succeed to unravel diseases hard-to-detect and easy-to-detect in an unsupervised manner: it is hard to detect local small lesions, such as brain abscess and enhanced lesions; but, it is easy to detect hyper-intense enhancing lesions, such as brain metastases (AUC $0.921$), especially for 7-SA MADGAN thanks to its feature recalibration. Our visualization of differences between real/reconstructed slices might play a key role in understanding and preventing various diseases, including rare disease.

Since we firstly propose a two-step unsupervised anomaly detection approach based on multiple slice reconstruction, its limitations are two-fold: yet less generalizable reconstruction and diagnosis.
As future work, we will investigate more suitable SA modules in a reconstruction model, such as Dual Attention Network that capture feature dependencies in both spatial/channel dimensions~\cite{fu2019dual}; here, optimizing where to place how many SA modules is the most relevant aspect. We will validate combining new loss functions for both reconstruction/diagnosis, including sparsity regularization~\cite{zhou2020sparse}, structural similarity~\cite{haselmann2018anomaly}, and perceptual loss~\cite{tuluptceva2020anomaly}. Lastly, we plan to collect a higher amount of healthy T1c scans to reliably detect and locate various diseases, including cancers and rare diseases. Integrating multi-modal imaging data, such as Positron Emission Tomography with specific radiotracers~\cite{rundoCMPB2017}, might further improve disease diagnosis~\cite{brier2016}, even when analyzed modalities are not always available~\cite{li2014multimodal}. Moreover, to specify detected anomalies, we might extend this work to supervised learning with limited pathological data by discriminating normal/pathological image distributions during diagnosis, instead of calculating the average $\ell _2$ loss per scan.


\begin{backmatter}

\section*{List of abbreviations used}
Area Under the Curve: AUCs, 
AutoEncoder: AE, 
Alzheimer's Disease: AD, 
Clinical Dementia Rating: CDR, 
Contrast-enhanced T1-weighted: T1c, 
Convolutional Neural Network: CNN, 
Computed Tomography: CT, 
Generative Adversarial Network: GAN, 
Magnetic Resonance Imaging: MRI, 
Medical Anomaly Detection Generative Adversarial Network: MADGAN, 
Mild Cognitive Impairment: MCI, 
Open Access Series of Imaging Studies-3: OASIS-3, 
Receiver Operating Characteristic: ROC, 
Rectified Linear Unit: ReLU, 
Self-Attention: SA, 
T1-weighted: T1, 
Variational AutoEncoder: VAE, 
Wasserstein loss with Gradient Penalty: WGAN-GP.

\section*{Competing interests}
  The authors declare that they have no competing interests.

\section*{Author's contributions}
Conceived the idea: CH, LR, ZAM, KM.
Designed the code: CH, LR, ZAM.
Collected the T1c dataset: TN.
Implemented the code: CH.
Performed the experiments: CH.
Analyzed the results: CH, LR.
Wrote the manuscript: CH, LR.
Critically read the manuscript and contributed to the discussion of the whole work: KM, TN, ZAM, YS, SK, ES, HN, SS.

\section*{Acknowledgements}
This research was partially supported both by AMED Grant Number JP18lk1010028 and The Mark Foundation for Cancer Research and Cancer Research UK Cambridge Centre [C9685/A25177].
Additional support has been provided by the National Institute of Health Research (NIHR) Cambridge Biomedical Research Centre.
Zolt\'{a}n \'{A}d\'{a}m Milacski was supported by Grant Number VEKOP-2.2.1-16-2017-00006. The OASIS-3 dataset has Grant Numbers P50 AG05681, P01 AG03991, R01 AG021910, P50 MH071616, U24 RR021382, and R01 MH56584.

\bibliographystyle{bmc-mathphys} 
\bibliography{bmc_article}      




\clearpage

\section*{Figures}
\begin{figure}[h]
  \centering
  \includegraphics[width=0.97\textwidth]{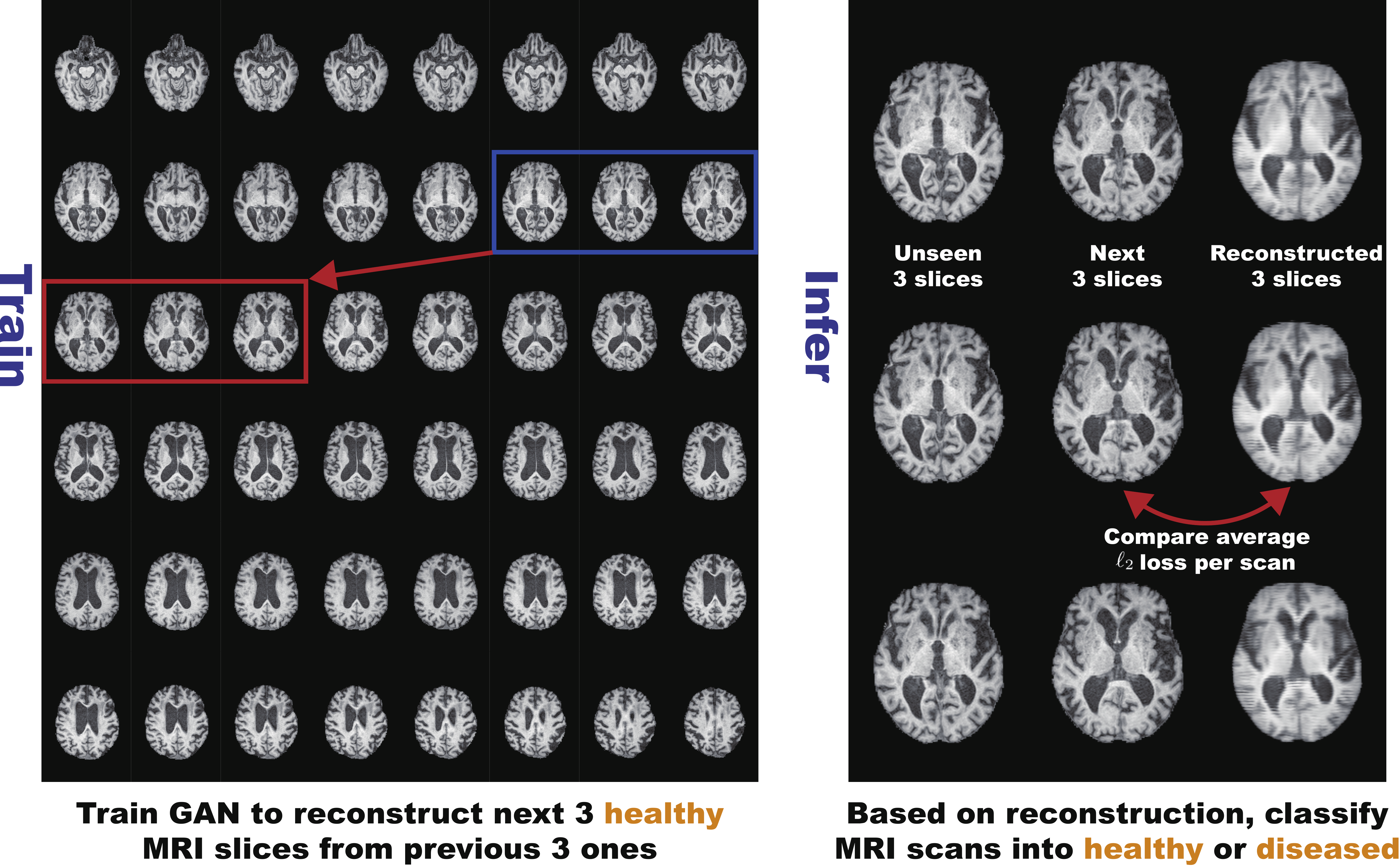}
\caption{Unsupervised medical anomaly detection framework: we train WGAN-GP w/ $\ell _1$ loss on $3$ healthy brain axial MRI slices to reconstruct the next $3$ ones, and test it on both unseen healthy and abnormal scans to classify them according to average $\ell _2$ loss per scan.}
\label{fig1}
\end{figure}

\begin{figure}[h]
  \centering
  \centerline{\includegraphics[width=0.97\textwidth]{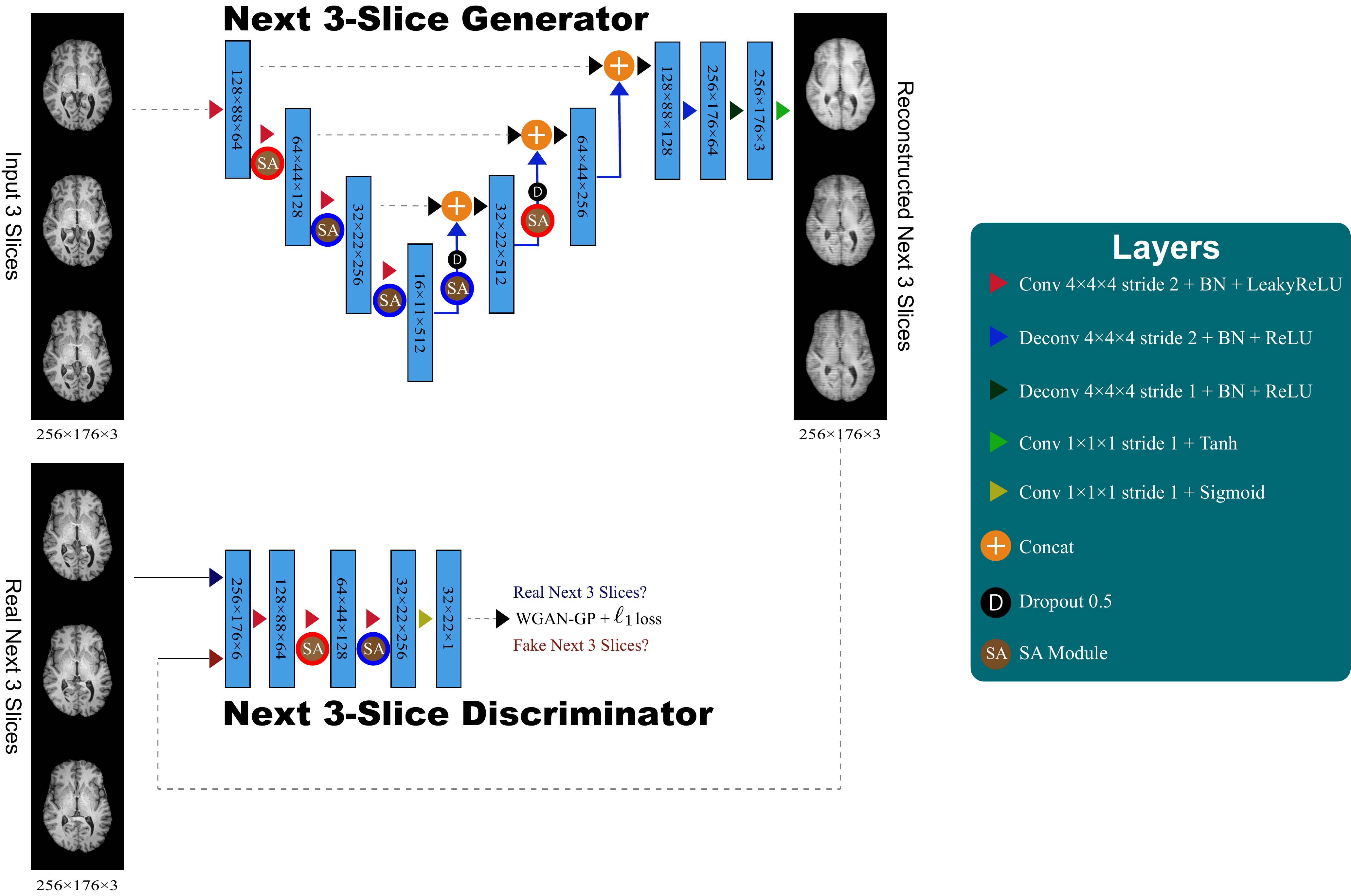}}
\caption{Proposed MADGAN architecture for the next $3$-slice generation from the input $3$ $256 \times 176$ brain MRI slices: 3-SA MADGAN has only 3 (red-contoured) SA modules after convolution/deconvolution whereas 7-SA MADGAN has 7 (red- and blue-contoured) SA modules. Similar to RGB images, we concatenate adjacent 3 gray slices into 3 channels.}
\label{fig2}
\end{figure}

\begin{figure}[h]
  \centering
  \includegraphics[width=0.97\columnwidth]{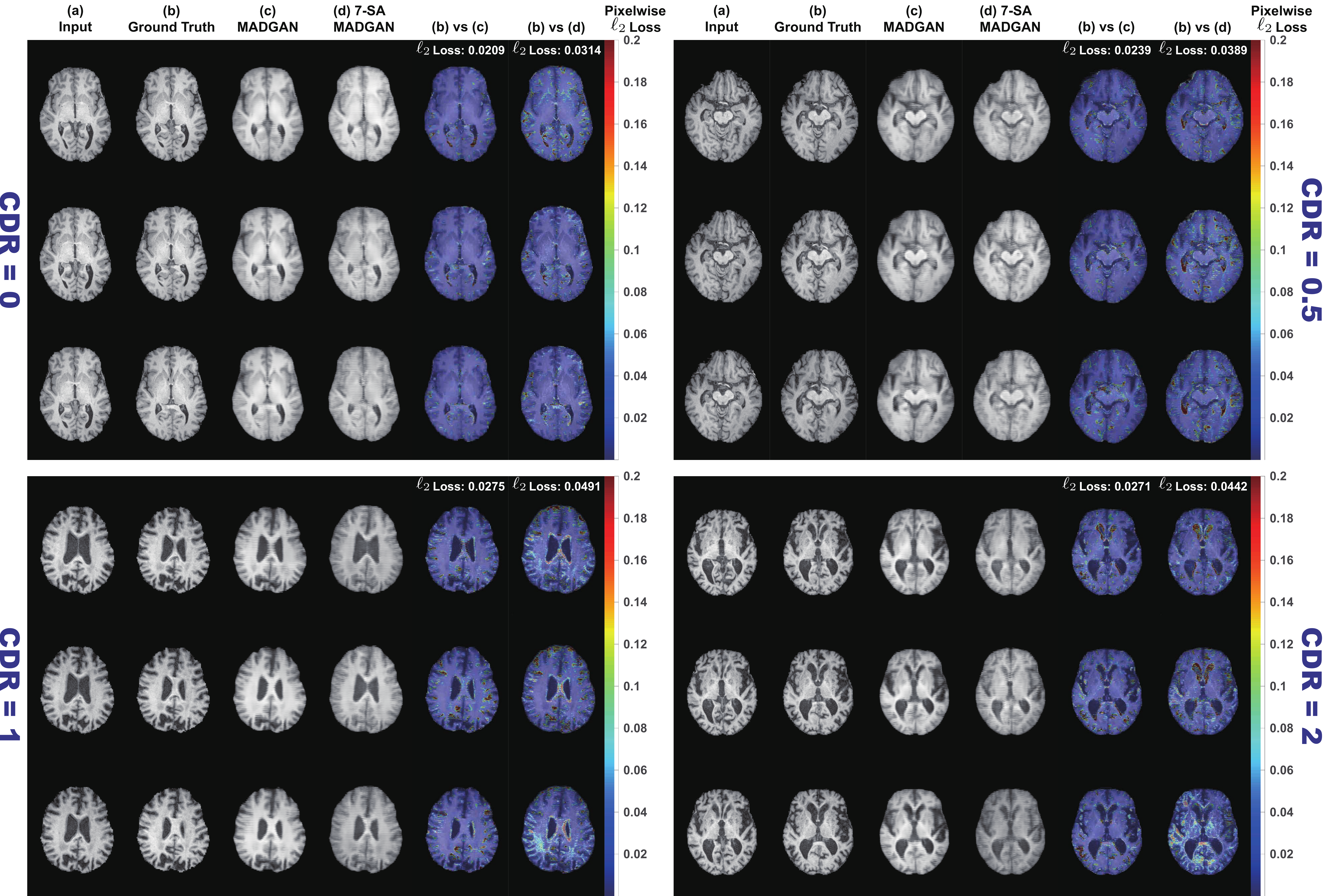}
\caption{Example T1 brain MRI slices with CDR $= 0/0.5/1/2$ from a test set: (\textbf{a}) Input $3$ real slices; (\textbf{b}) Ground truth next $3$ real slices; (\textbf{c}, \textbf{d}) Next $3$ slices reconstructed by MADGAN and 7-SA MADGAN. To compare the real/reconstructed next $3$ slices, we show pixelwise $\ell _2$ loss values in (\textbf{b}) vs (\textbf{c}) and (\textbf{b}) vs (\textbf{d}) columns, respectively. Using a Jet colormap in $[0, 0.2]$ with alpha-blending, we overlay the obtained maps onto the ground truth slices. The achieved 
slice-level, pixelwise $\ell _2$ loss values are also displayed.}
\label{fig3}
\end{figure}

\begin{figure}[h]
  \centering
  \includegraphics[width=0.97\columnwidth]{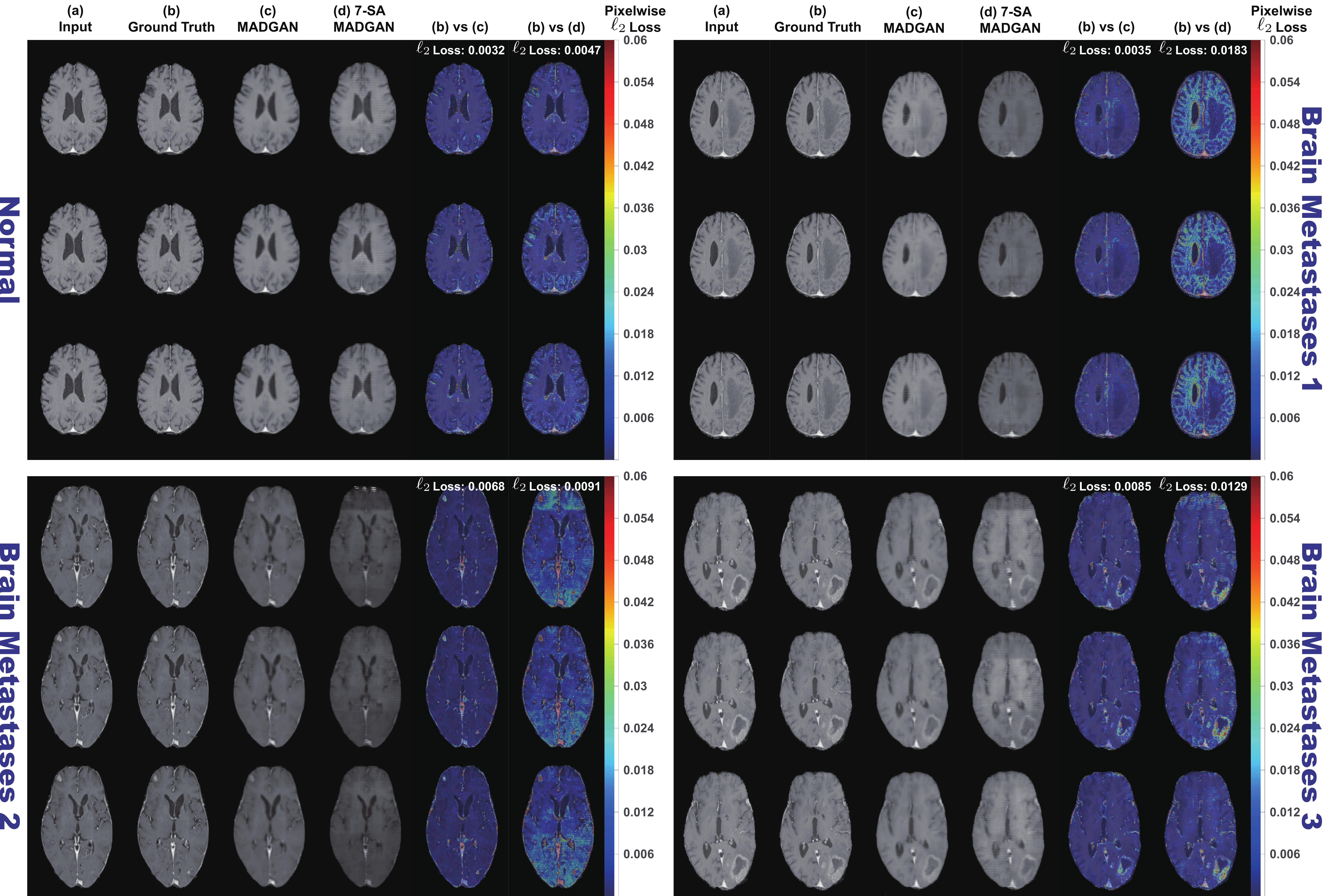}
\caption{Example T1c brain MRI slices with no abnormal findings/three brain metastases from a test set: (\textbf{a}) Input $3$ real slices; (\textbf{b}) Ground truth next $3$ real slices; (\textbf{c}, \textbf{d}) Next $3$ slices reconstructed by MADGAN and 7-SA MADGAN. To compare the real/reconstructed next $3$ slices, we show pixelwise $\ell _2$ loss values in (\textbf{b}) vs (\textbf{c}) and (\textbf{b}) vs (\textbf{d}) columns, respectively. Using a Jet colormap in $[0, 0.06]$ with alpha-blending, we overlay the obtained maps onto the ground truth slices. The achieved 
slice-level, pixelwise $\ell _2$ loss values are also displayed.}
\label{fig4}
\end{figure}

\begin{figure}[h]
  \centering
  \includegraphics[width=0.97\columnwidth]{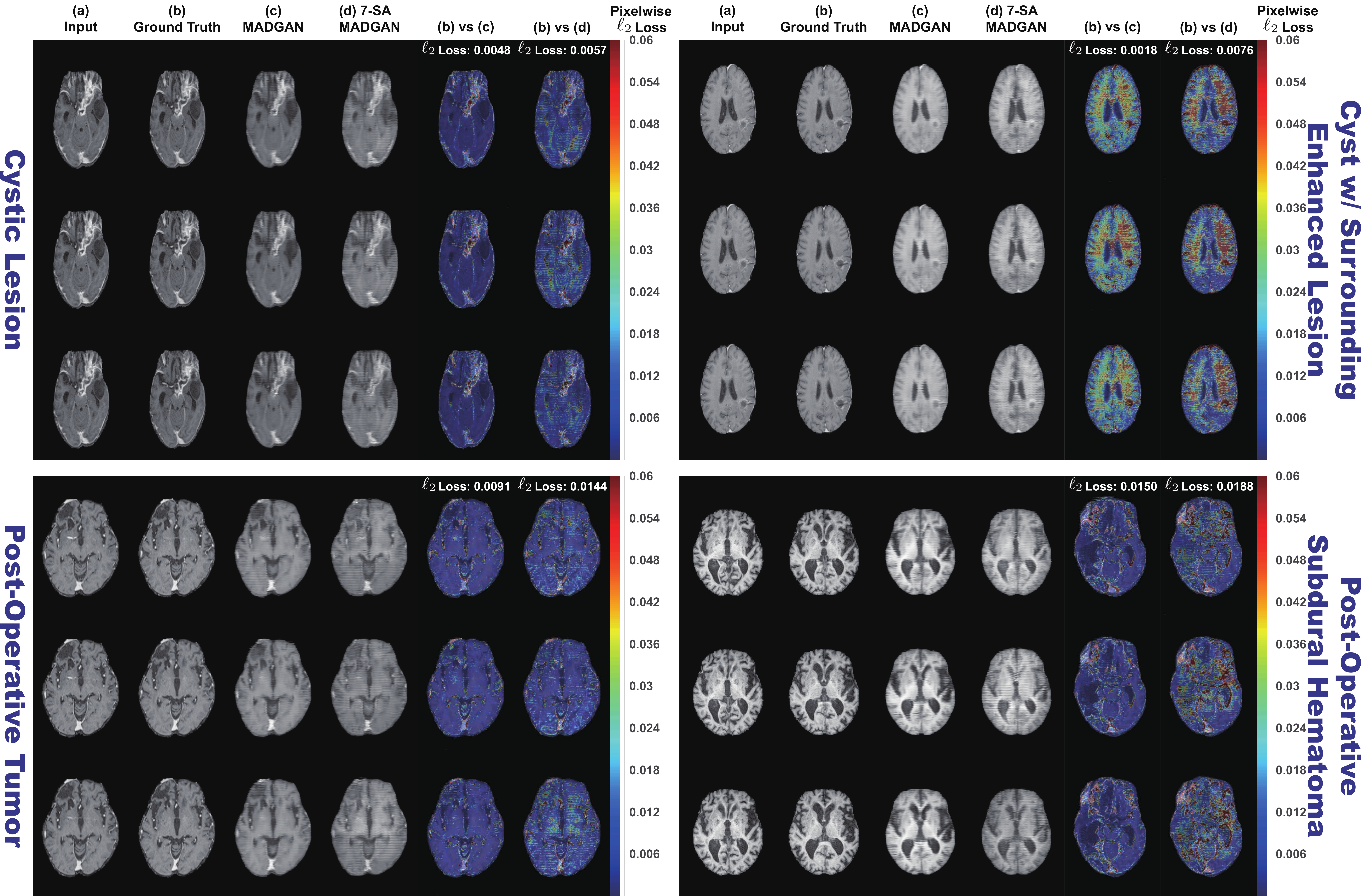}
\caption{Example T1c brain MRI slices with four different brain diseases from a test set: (\textbf{a}) Input $3$ real slices; (\textbf{b}) Ground truth next $3$ real slices; (\textbf{c}, \textbf{d}) Next $3$ slices reconstructed by MADGAN and 7-SA MADGAN. To compare the real/reconstructed next $3$ slices, we show pixelwise $\ell _2$ loss values in (\textbf{b}) vs (\textbf{c}) and (\textbf{b}) vs (\textbf{d}) columns, respectively. Using a Jet colormap in $[0, 0.06]$ with alpha-blending, we overlay the obtained maps onto the ground truth slices. The achieved 
slice-level, pixelwise $\ell _2$ loss values are also displayed.}
\label{fig5}
\end{figure}

\begin{figure}[h]
  \centering
  \includegraphics[width=0.97\columnwidth]{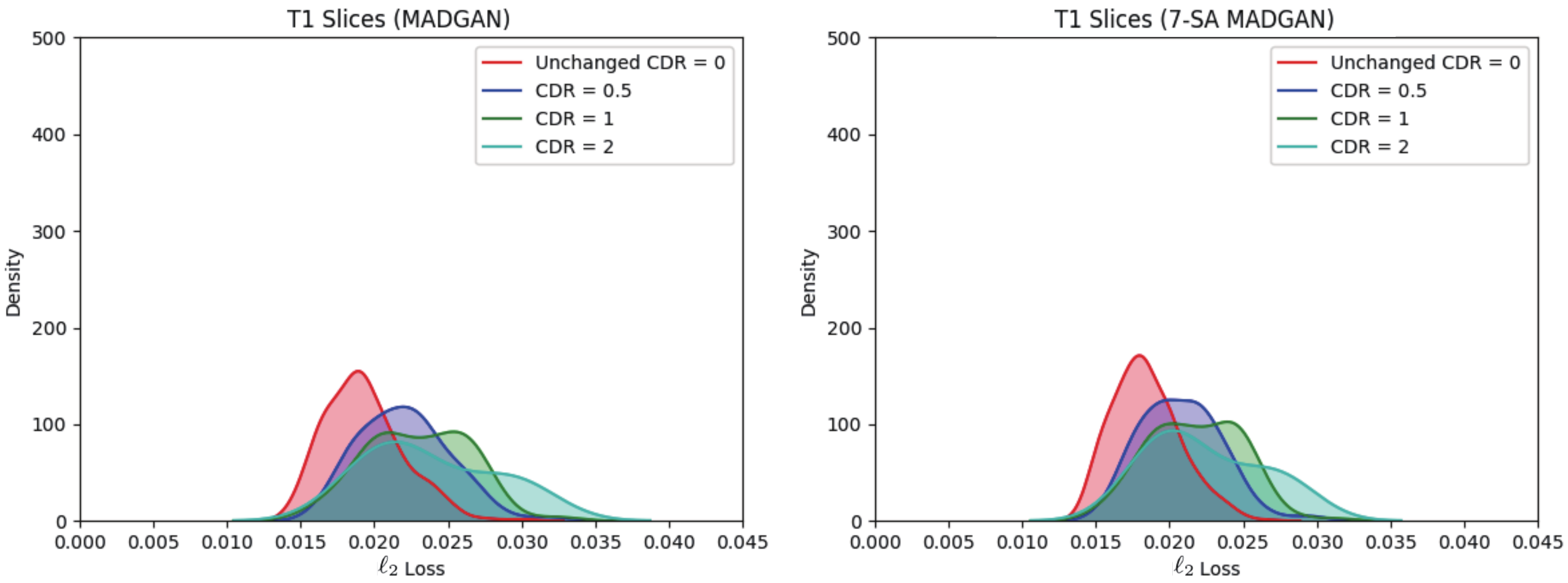}
\caption{Distributions of average $\ell _2$ loss per scan evaluated on T1 slices with CDR $= 0/0.5/1/2$ reconstructed by: (\textbf{a}) MADGAN and (\textbf{b}) 7-SA MADGAN.}
\label{fig6}
\end{figure}

\begin{figure}[h]
  \centering
  \includegraphics[width=0.97\columnwidth]{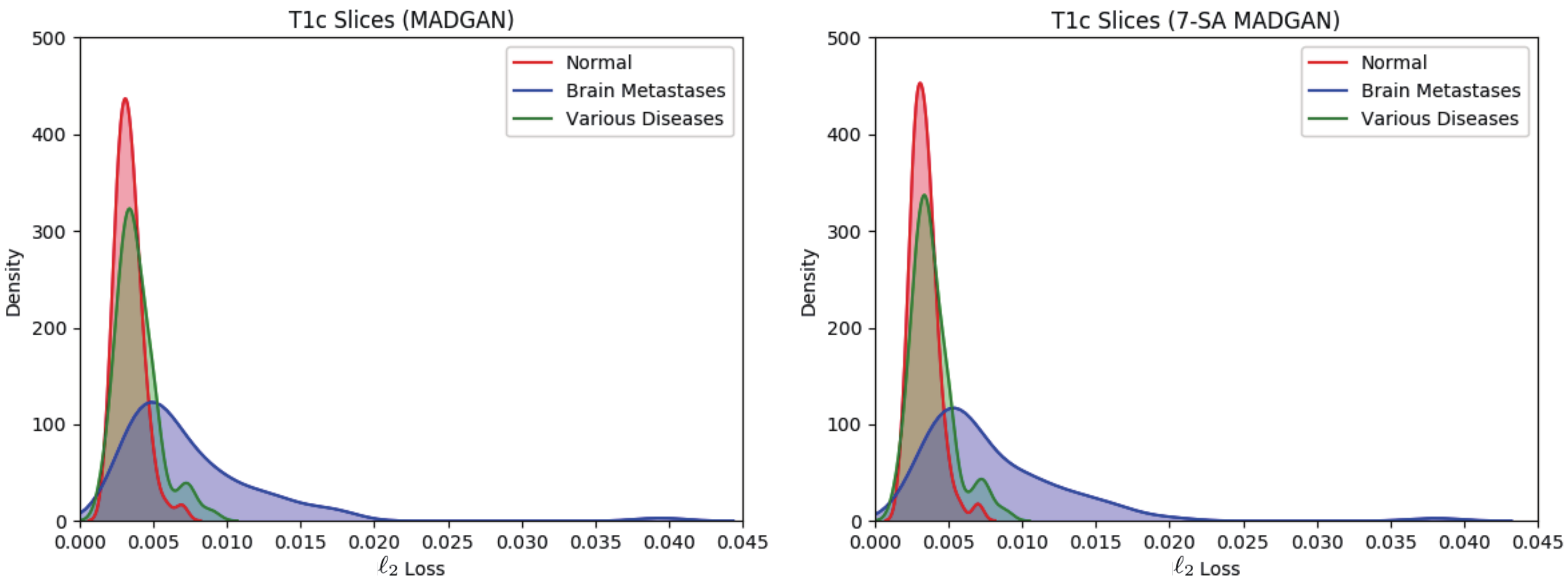}
\caption{Distributions of average $\ell _2$ loss per scan evaluated on T1c slices with no abnormal findings/brain metastases/various diseases reconstructed by: (\textbf{a}) MADGAN and (\textbf{b}) 7-SA MADGAN.}
\label{fig7}
\end{figure}

\begin{figure}[h]
  \centering
  \includegraphics[width=0.97\columnwidth]{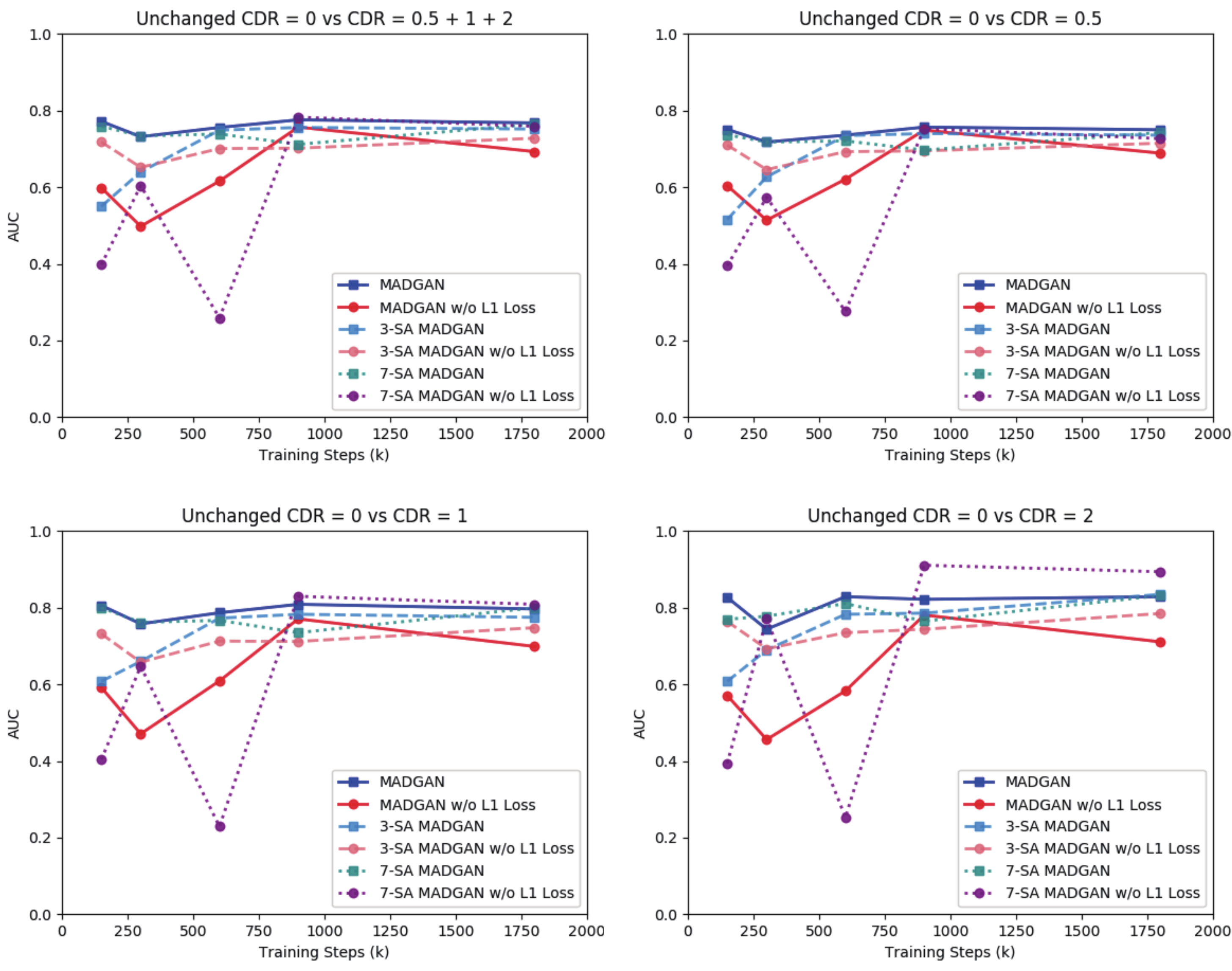}
\caption{AUC performance on T1 scans using average $\ell _2$ loss per scan under different training steps (i.e., $150$k, $300$k, $600$k, $900$k, $1.8$M steps). Unchanged CDR $= 0$ (i.e., cognitively healthy population) is compared against: (\textbf{a}) all the other CDRs (i.e., dementia); (\textbf{b}) CDR $= 0.5$ (i.e., very mild dementia); (\textbf{c}) CDR $= 1$ (i.e., mild dementia); (\textbf{d}) CDR $= 2$ (i.e., moderate dementia).}
\label{fig8}
\end{figure}


\begin{figure}[h]
  \centering
  \includegraphics[width=0.97\columnwidth]{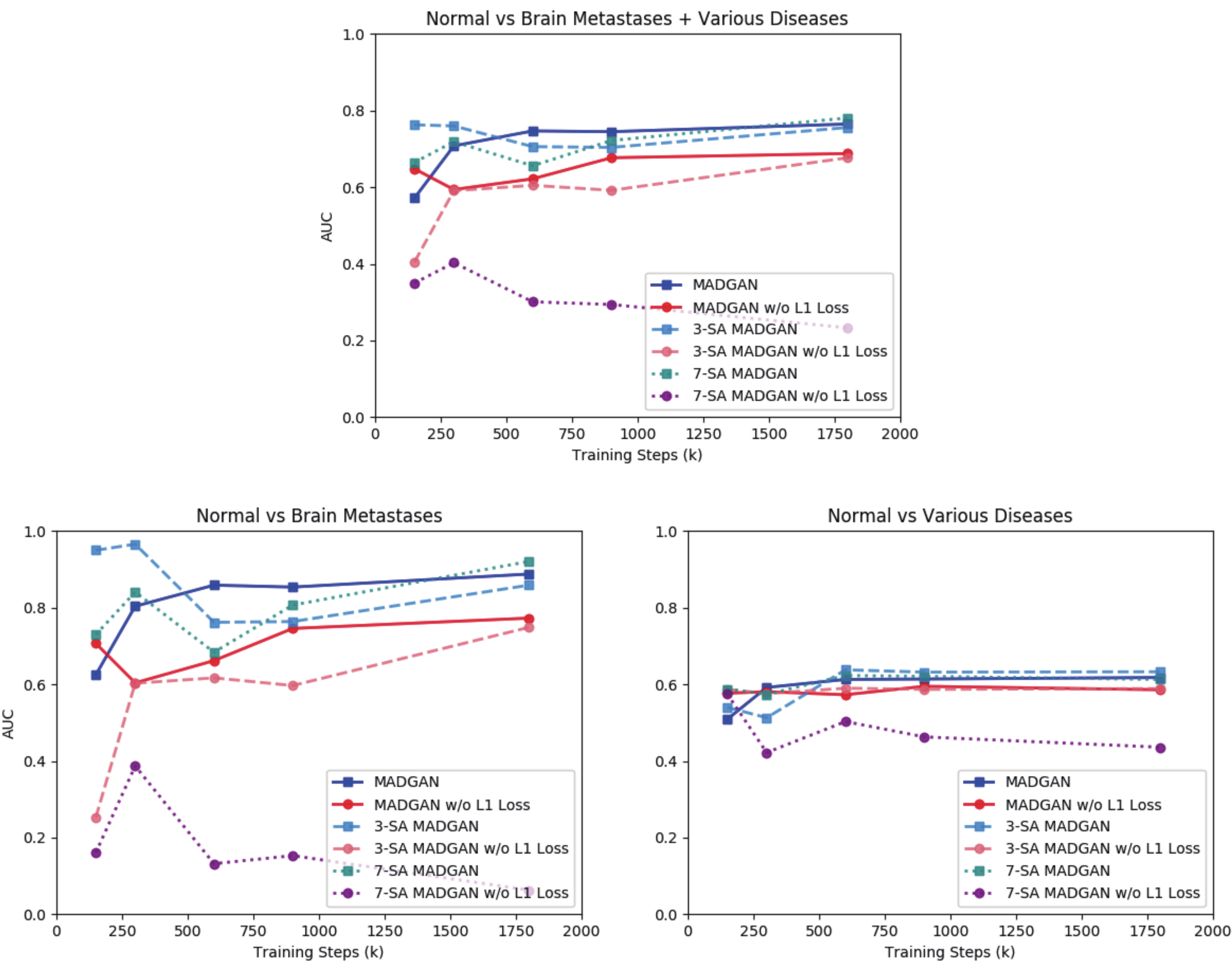}
\caption{AUC performance on T1c scans using average $\ell _2$ loss per scan under different training steps (i.e., $150$k, $300$k, $600$k, $900$k, $1.8$M steps). No abnormal findings are compared against: (\textbf{a}) brain metastases + various diseases; (\textbf{b}) brain metastases; (\textbf{c}) various diseases.}
\label{fig9}
\end{figure}

\begin{figure}[h]
  \centering
  \includegraphics[width=0.97\columnwidth]{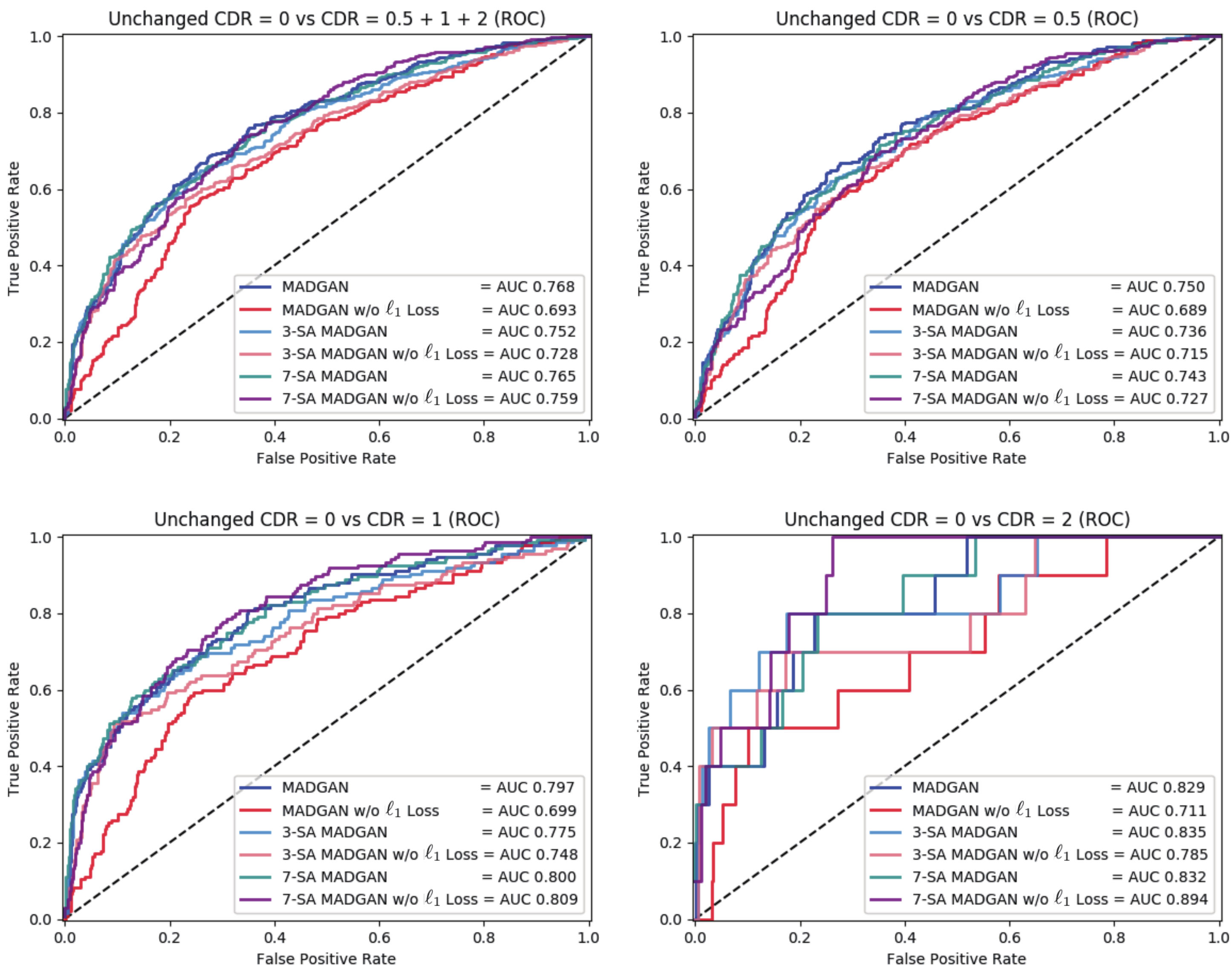}
\caption{Unsupervised anomaly detection results using average $\ell _2$ loss per scan on reconstructed T1 slices (ROCs and AUCs). Unchanged CDR $= 0$ (i.e., cognitively healthy population) is compared against:  (\textbf{a}) all the other CDRs (i.e., dementia); (\textbf{b}) CDR $= 0.5$ (i.e., very mild dementia); (\textbf{c}) CDR $= 1$ (i.e., mild dementia); (\textbf{d}) CDR $= 2$ (i.e., moderate dementia). Each model is trained for 1.8M steps.}
\label{fig10}
\end{figure}

\begin{figure}[h]
  \centering
  \includegraphics[width=0.97\columnwidth]{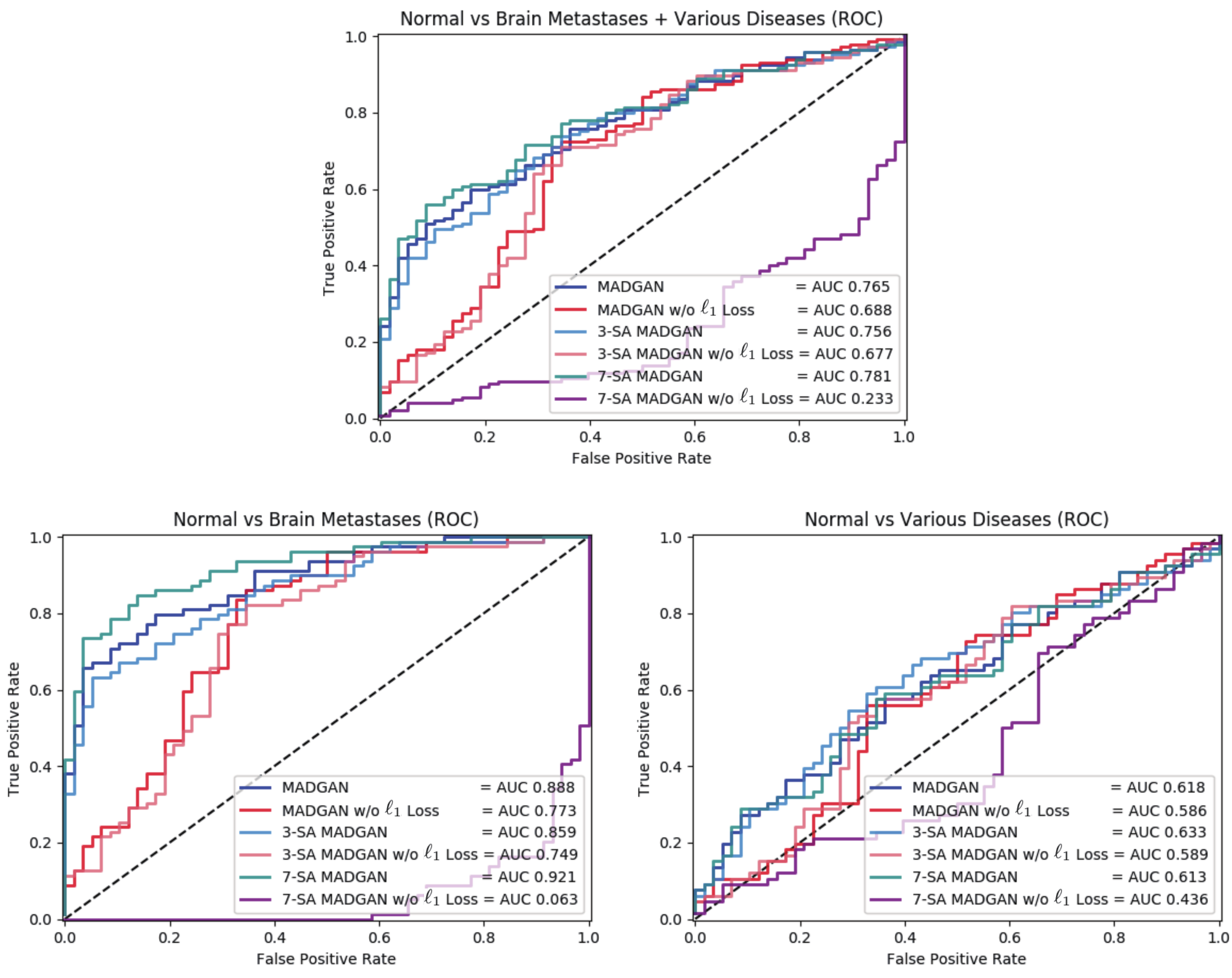}
\caption{Unsupervised anomaly detection results using average $\ell _2$ loss per scan on reconstructed T1c slices (ROCs and AUCs). No abnormal findings are compared against: (\textbf{a}) brain metastases + various diseases; (\textbf{b}) brain metastases; (\textbf{c}) various diseases. Each model is trained for 1.8M steps.}
\label{fig11}
\end{figure}


\clearpage
\begin{table}[h]
\caption{AUC performance of unsupervised anomaly detection on T1 scans using average $\ell _2$ loss (among whole slice sets/continuous 10 slice sets exhibiting the highest loss) per scan. Unchanged CDR $= 0$ (i.e., cognitively healthy population) is compared against: Unchanged CDR $= 0$ (i.e., cognitively healthy population) is compared against:  (\textit{i}) all the other CDRs (i.e., dementia); (\textit{ii}) CDR $= 0.5$ (i.e., very mild dementia); (\textit{iii}) CDR $= 1$ (i.e., mild dementia); (\textit{iv}) CDR $= 2$ (i.e., moderate dementia). Each model is trained 
for 1.8M steps.}
\centering
\begin{small}
\scalebox{0.84}{
\begin{tabular}{lrrrr}
\Hline\noalign{\smallskip}
\bfseries CDR = 0 vs & \bfseries  CDR = 0.5 + 1 + 2 & \bfseries CDR = 0.5 & \bfseries CDR = 1 & 
\bfseries CDR = 2 \\
\noalign{\smallskip}\hline\noalign{\smallskip}

MADGAN & \textbf{0.768} & \textbf{0.750} & 0.797 & 0.829\\
MADGAN (10 slice sets) & 0.764 & 0.745 &	0.793 &	0.830\\
\noalign{\smallskip}\hline\noalign{\smallskip}
MADGAN w/o $\ell _1$ Loss  &	0.693 &	0.689 &	0.699 &	0.711\\
MADGAN w/o $\ell _1$ Loss (10 slice sets) &	0.705 &	0.697 &	0.717 &	0.736\\
\noalign{\smallskip}\hline\noalign{\smallskip}
3-SA MADGAN &	0.752 &	0.736 &	0.775 &	0.835\\
3-SA MADGAN (10 slice sets) & 0.739 &0.725 & 0.760 & 0.810\\
\noalign{\smallskip}\hline\noalign{\smallskip}
3-SA MADGAN w/o $\ell _1$ Loss & 0.728 & 0.715 & 0.748 & 0.785\\
3-SA MADGAN w/o $\ell _1$ Loss (10 slice sets) & 0.735 & 0.721 & 0.756 & 0.806\\
\noalign{\smallskip}\hline\noalign{\smallskip}

7-SA MADGAN & 0.765 & 0.743 & 0.800 & 0.832\\
7-SA MADGAN (10 slice sets) & 0.764 & 0.743 & 0.798 & 0.835\\
\noalign{\smallskip}\hline\noalign{\smallskip}

7-SA MADGAN w/o $\ell _1$ Loss & 0.759 & 0.727 & \textbf{0.809} & \textbf{0.894}\\
7-SA MADGAN w/o $\ell _1$ Loss (10 slice sets) & 0.746 & 0.710 & 0.803 &0.868\\
\noalign{\smallskip}\Hline\noalign{\smallskip}
\end{tabular}}
\label{tab:CDR}
\end{small}
\end{table}

\begin{table}[h]
\caption{AUC performance of unsupervised anomaly detection on T1c scans using average $\ell _2$ loss (among whole slice sets/continuous 10 slice sets exhibiting the highest loss) per scan. No abnormal findings are compared against: (\textit{i}) brain metastases + various diseases; (\textit{ii}) brain metastases; (\textit{iii}) various diseases. Each model is trained 
for 1.8M steps.}
\centering
\begin{small}
\begin{tabular}{lrrr}
\Hline\noalign{\smallskip}
\bfseries Normal vs & \bfseries  BM + VD & \bfseries BM & \bfseries VD\\
\noalign{\smallskip}\hline\noalign{\smallskip}

MADGAN & 0.765 &	0.888 & 0.618
\\
MADGAN (10 slice sets) & 0.769 & 0.905 & 0.607\\
\noalign{\smallskip}\hline\noalign{\smallskip}
MADGAN w/o $\ell _1$ Loss &	0.688 & 0.773 & 0.586\\
MADGAN w/o $\ell _1$ Loss (10 slice sets) &	0.696 & 0.778 &0.597\\
\noalign{\smallskip}\hline\noalign{\smallskip}
3-SA MADGAN  &	0.756 & 0.859 & \textbf{0.633}\\
3-SA MADGAN (10 slice sets) & 0.760 & 0.871 & 0.626\\
\noalign{\smallskip}\hline\noalign{\smallskip}
3-SA MADGAN w/o $\ell _1$ Loss & 0.677 & 0.749 & 0.589\\
3-SA MADGAN w/o $\ell _1$ Loss (10 slice sets) & 0.708 & 0.780 & 0.622\\
\noalign{\smallskip}\hline\noalign{\smallskip}

7-SA MADGAN  & \textbf{0.781} & \textbf{0.921} & 0.613\\
7-SA MADGAN (10 slice sets) & 0.776 & 0.917 & 0.608\\
\noalign{\smallskip}\hline\noalign{\smallskip}

7-SA MADGAN w/o $\ell _1$ Loss  & 0.233 & 0.063 & 0.436\\
7-SA MADGAN w/o $\ell _1$ Loss  (10 slice sets) & 0.234 & 0.091 & 0.405\\
\noalign{\smallskip}\Hline\noalign{\smallskip}

\label{tab:T1c}
\end{tabular}
\end{small}
\end{table}

\end{backmatter}
\end{document}